\documentclass[10pt,twocolumn,letterpaper]{article}

\usepackage[pagenumbers]{iccv} % To force page numbers, e.g. for an arXiv version

\usepackage{multirow} 
\usepackage{array}
\usepackage{graphicx}  % 处理旋转文本
\usepackage{booktabs}  % 处理表格
\usepackage{rotating}  % 旋转文本

\definecolor{iccvblue}{rgb}{0.21,0.49,0.74}
\usepackage[pagebackref,breaklinks,colorlinks,allcolors=iccvblue]{hyperref}

\title{Towards Privacy-Preserving Fine-Grained Visual Classification \\ via Hierarchical Learning from Label Proportions}

\author{Jinyi Chang$^1$
    \ \ Dongliang Chang$^{1 \dag}$
    \ \ Lei Chen$^2$
    \ \ Bingyao Yu$^2$
    \ \ Zhanyu Ma$^1$
\vspace{0.3em}
\\\small $^1$PRIS, Beijing University of Posts and Telecommunications
\\\small $^2$Tsinghua University
}

\begin{document}
\maketitle
\makeatletter\def\Hy@Warning#1{}\makeatother
\let\thefootnote\relax\footnotetext{$\dag$ Corresponding author}
\begin{abstract}
In recent years, Fine-Grained Visual Classification (FGVC) has achieved impressive recognition accuracy, despite minimal inter-class variations. However, existing methods heavily rely on instance-level labels, making them impractical in privacy-sensitive scenarios such as medical image analysis. This paper aims to enable accurate fine-grained recognition without direct access to instance labels. To achieve this, we leverage the Learning from Label Proportions (LLP) paradigm, which requires only bag-level labels for efficient training. Unlike existing LLP-based methods, our framework explicitly exploits the hierarchical nature of fine-grained datasets, enabling progressive feature granularity refinement and improving classification accuracy. We propose \textbf{Learning from Hierarchical Fine-Grained Label Proportions (LHFGLP)}, a framework that incorporates Unrolled Hierarchical Fine-Grained Sparse Dictionary Learning, transforming handcrafted iterative approximation into learnable network optimization. Additionally, our proposed Hierarchical Proportion Loss provides hierarchical supervision, further enhancing classification performance. Experiments on three widely-used  fine-grained datasets, structured in a bag-based manner, demonstrate that our framework consistently outperforms existing LLP-based methods. We will release our code and datasets to foster further research in privacy-preserving fine-grained classification.
% and endangered species identification
\end{abstract}  
\section{Introduction}
\label{sec:intro}
\begin{figure}[t]
  \centering
  % \begin{subfigure}{0.99\columnwidth}
  %   \includegraphics[width=0.98\textwidth]{models/model1-1.pdf} 
  %   \caption{General Visual Classification.}
  %   \label{fig:a}
  % \end{subfigure}
  % \begin{subfigure}{0.99\columnwidth}
  %   \includegraphics[width=0.98\textwidth]{models/model1-2.pdf}
  %   \caption{LLP for General Visual Classification.}
  %   \label{fig:b}
  % \end{subfigure}
  % \begin{subfigure}{0.99\columnwidth}
  %   \includegraphics[width=0.98\textwidth]{models/model1-3.pdf}
  %   \caption{LLP for Fine-Grained Visual Classification.}
  %   \label{fig:c}
  % \end{subfigure}
  \begin{subfigure}{0.98\columnwidth}
    \includegraphics[width=0.98\textwidth]{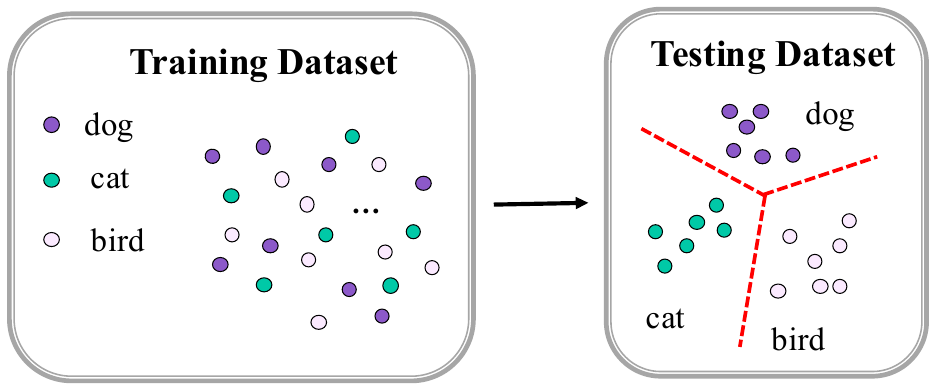} 
    \caption{General image classification.}
    \label{fig:a}
  \end{subfigure}
  \begin{subfigure}{0.98\columnwidth}
    \includegraphics[width=0.98\textwidth]{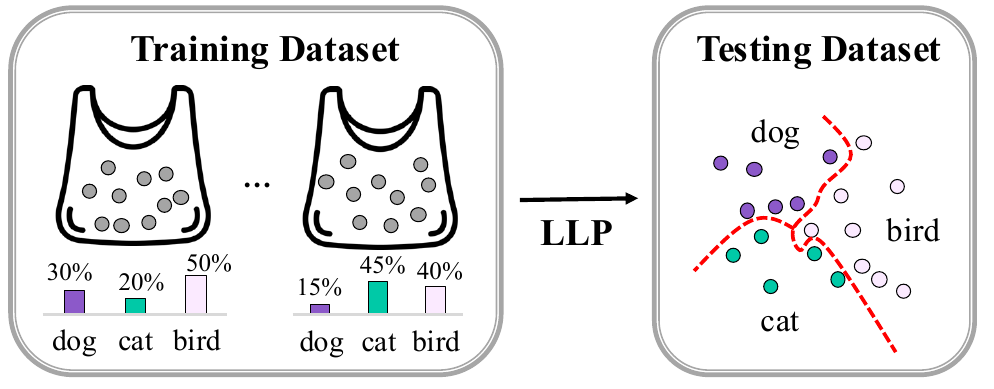}
    \caption{Learning from Label Proportions.}
    \label{fig:b}
  \end{subfigure}
  \caption{
  Comparison of general image classification and Learning from Label Proportions (LLP). Individual images are represented as dots, colored by their ground-truth categories. Under the LLP paradigm, images are aggregated into bags with only bag-level label proportions, which are in grey to indicate unknown categories. The red dashed lines reflect the efforts required for image classification.
  % Each point represents one image with the color indicates its category label. Bags of grey dots denote images whose categories are unknown, but are packed and annotated with bag-level label proportions. The red dashed lines indicate the efforts required for image classification.
  }
  \label{model1}
\end{figure}

% Fine-Grained Visual Classification (FGVC) has become a pivotal task in computer vision, aiming to distinguish highly similar subcategories, such as bird species, medical lesion subtypes, or vehicle models, through capturing subtle differences in specific regions of images. With the progress of deep learning, FGVC has achieved impressive accuracy despite the challenges of small inter-class differences and large intra-class variations. Current dominant approaches typically rely on extensive instance-level labels to guide the localization of discriminative regions, enabling models to progressively refine the recognition of features from global shapes to fine-grained details \cite{ke2023granularity,chen2024fet,sikdar2024interweaving}. While these methods exhibit excellent recognition performance, their over-reliance on precise instance-level annotations poses a critical and yet under-explored challenge, the risk of privacy leakage in sensitive applications.

Fine-Grained Visual Classification (FGVC) aims to distinguish visually similar subcategories, such as bird species, medical lesions, or vehicle models, by capturing subtle yet discriminative visual patterns. Recent advances in deep learning have significantly improved FGVC performance through techniques such as hierarchical feature fusion \cite{du2020fine}, cross-layer self-distillation \cite{ke2023granularity}, and graph-based global-local interactions \cite{chen2024fet,sikdar2024interweaving}, effectively mitigating challenges posed by small inter-class differences and large intra-class variations.

Despite these advancements, existing FGVC methods rely heavily on instance-level labels, making them impractical for privacy-sensitive applications, such as medical image analysis, where instance labels, despite being available, are often protected by strict privacy regulations. This results in a fundamental bottleneck: while high-quality labeled data exists, models are unable to access it for training, hindering the application of FGVC in domains requiring privacy preservation. Addressing this challenge requires a paradigm shift towards learning fine-grained representations without direct access to instance-level labels.

Learning from Label Proportions (LLP) offers a promising solution by enabling model training without instance-level supervision. Instead of requiring explicit instance annotations, LLP operates on bag-level labels, which specify only the category distribution within a group of instances. As illustrated in \cref{model1}, data owners, constrained by privacy policies, can randomly aggregate instances into bags and provide corresponding bag labels, which indicate the proportion of different classes within each bag (e.g., \{0.3: dog, 0.2: cat, 0.5: bird\} means that 30\% of the instances are dogs, 20\% are cats and 50\% are birds). Since bag labels do not reveal individual instance identities, this approach inherently preserves data privacy. LLP algorithms leverage multiple such bags to facilitate effective feature learning under weak supervision, and several LLP-based methods have been explored in coarse-grained classification tasks \cite{tsai2020learning,zhang2022learning,asanomi2023mixbag}.

However, directly applying existing LLP algorithms to FGVC is non-trivial due to the unique challenge of fine-grained feature extraction. 
Training purely with bag-level supervision is expected to lead to induce significant performance degradation, often inferior to the baselines trained with instance-level supervision. 
Most LLP algorithms assume relatively distinct inter-class variations and lack explicit mechanisms to capture the subtle intra-class distinctions required for FGVC. 
% As a result, models trained purely with bag-level supervision struggle to learn discriminative fine-grained features, leading to significant performance degradation, often inferior to a naïve baseline trained with limited supervision. 
As a result, these models struggle to learn discriminative fine-grained features under the bag-level supervision, leading to significant performance degradation.
% , often inferior to a naïve baseline trained with limited supervision. 
Thus, a fundamental challenge remains: how can LLP be adapted to effectively learn fine-grained representations without instance-level labels?

To address this issue, we propose \textbf{Learning from Hierarchical Fine-Grained Label Proportions (LHFGLP)}, a novel framework that synergizes LLP with the hierarchical structure inherent in fine-grained data. Unlike conventional LLP approaches, LHFGLP explicitly aligns bag-level supervision with fine-grained feature learning, enabling progressive refinement of feature granularity through a hierarchical learning process. This strategy is inspired by human perception, where fine-grained recognition naturally progresses from coarse to fine-grained details. Specifically, we introduce an \textbf{Unrolled Hierarchical Fine-Grained Sparse Dictionary Learning} strategy, which reformulates handcrafted iterative approximation into an end-to-end learnable network optimization process via deep unrolling. By learning an over-complete dictionary, fine-grained features are represented as linear combinations of feature atoms, enabling a more flexible and precise feature representation. Additionally, imposing sparsity constraints forces the model to focus on the most relevant and discriminative regions, thereby improving robustness. Furthermore, we design a \textbf{Hierarchical Proportion Loss}, which guides the progressive refinement of fine-grained features, improving the model’s ability to extract discriminative details under bag-level supervision.

Our contributions can be summarized as follows:
\begin{itemize}
    \item We introduce a LLP-based framework for privacy-preserving FGVC, eliminating the reliance on instance-level labels during training. By leveraging only bag-level annotations, our approach ensures privacy protection at the source of data collection, mitigating the risk of sensitive information leakage in FGVC applications. To the best of our knowledge, this is the first attempt at privacy-preserving FGVC based on LLP.
    \item We propose LHFGLP, an LLP-based learning framework dedicated to FGVC, integrating Unrolled Hierarchical Fine-Grained Sparse Dictionary Learning with Hierarchical Proportional Loss to facilitate fine-grained feature extraction.
    % with progressively refined granularity. 
    Additionally, LHFGLP is modular and enables seamless plug-and-play integration into existing FGVC pipelines for privacy-sensitive applications.
    % Additionally, LHFGLP is modular and can be seamlessly integrated into existing FGVC pipelines in a plug-and-play fashion for privacy-sensitive applications.
    \item We conduct extensive experiments on three widely used FGVC datasets structured in a bag-based manner for evaluation. Quantitative results, ablation studies, and visualization analysis demonstrate that LHFGLP consistently outperforms existing LLP-based architectures, showcasing its effectiveness in privacy-preserving FGVC.
\end{itemize}

This paper provides a novel perspective on privacy-conscious fine-grained visual classification, demonstrating that accurate fine-grained classification can be achieved without direct access to instance-level labels. We will publicly release our code and datasets to facilitate further research in privacy-preserving fine-grained classification.

\begin{figure*}[t]
\centering
\includegraphics[width=0.94\textwidth]{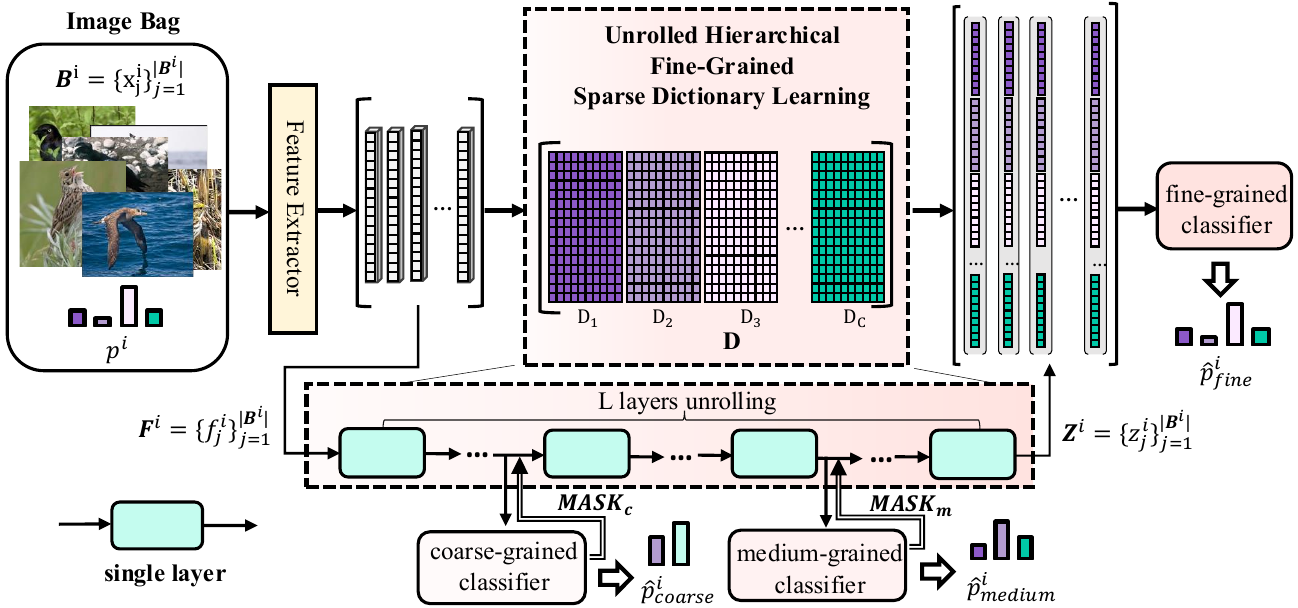} % Reduce the figure size so that it is slightly narrower than the column.
\caption{Overview of our proposed LHFGLP framework, which could integrate our Unrolled Hierarchical Fine-Grained Sparse Dictionary Learning with the fundamental FGVC pipeline in a plug-and-play manner.}
\label{model3}
\end{figure*}
%-------------------------------------------------------------------------
\section{Related Works}
\label{sec:related}

%-------------------------------------------------------------------------

\subsection{Fine-Grained Visual Classification (FGVC)}
% Current research in FGVC is mostly oriented towards tasks such as image classification with image level category labels, where the core idea is to extract more discriminative features in response to the small inter-class differences and large intra-class variations among fine-grained images \cite{du2020fine, gwilliam2021fair, he2022transfg,ke2023granularity, pan2023ssfe, chen2024fet, xu2024context}. 
% PMG utilized fine-grained hierarchical structures to learn at different granularities, and then progressively fused at multiple granularities to facilitate feature learning \cite{du2020fine}. Considering the diversity of local discriminative semantics and contextual structural associations, \citeauthor{ke2023granularity} proposed GDSMP-Net, which introduced granularity-aware distillation to discover subtle differences at each granularity and cross-layer self-distillation regularization to improve the robustness across different granularities \cite{ke2023granularity}. CSQA-Net emphasized the interactions between multi-scale part features and global semantics to model contextual relationships between different parts, thus extracting more discriminative fine-grained features\cite{xu2024context}.
% However, labeling fine-grained annotations is a tedious and expensive manual task and is highly sensitive to human annotation errors.
Deep learning as a powerful tool has made significant breakthroughs in computer vision tasks. In contrast to general image recognition, FGVC requires the model to focus on fine-grained regions, which are often difficult to quickly localize and identify. Early FGVC approaches relied heavily on additional dense annotations to help target these regions \cite{wei2018mask}, and gradually shifted to only requiring image labels. 
% Relevant research include unsupervised detection or segmentation, and attention mechanisms to guide discriminative feature extractions \cite{yang2018learning,chang2020devil}.
Extracting features with small inter-class differences and large intra-class variations among fine-grained images is the core concept of FGVC \cite{du2020fine, gwilliam2021fair, he2022transfg,pan2023ssfe, chen2024fet, sikdar2024interweaving}. FGoN utilizes level-specific classification heads to distinguish between different levels of features, and exploits the finer-grained features to participate in coarser-grained label predictions \cite{chang2021your}. LHT uses a probabilistic framework to tackle hierarchical classification \cite{wang2021label}. PMG facilitates fine-grained feature learning via the fine-grained hierarchical structure at different granularities with progressive feature fusion \cite{du2020fine}. 
GDSMP-Net explores subtle differences at each granularity by introducing a granularity-aware distillation and improves the robustness through a cross-layer self-distillation regularization \cite{ke2023granularity}. 
The CSQA-Net emphasises the interaction between multi-scale part features and global semantics, trying to extract more discriminative fine-grained features by simulating the contextual relationships among different parts \cite{xu2024context}. 
% CAFL encodes the prediction consistency constraint into a weak supervision mechanism through forward deduction and backward induction over the fine-grained label hierarchy, with a disentanglement and bidirectional reinforcement classification head to extract features \cite{wang2023consistency}. 
MGCF turns inter-category similarity relations into fine-grained recognition performance, exploring associations between label hierarchies in multi-granularity prediction \cite{shu2023fine}.
\subsection{Learning from Label Proportions (LLP)}
% Like other learning tasks, LLP was initially attempted to be solved by SVM \cite{yu2013proptosvm} or logistic hypothesis \cite{yu2014learning}. With the development of deep learning, researchers are gradually trying to use deep learning for LLP tasks. Batch Averager, proposed by \citeauthor{ardehaly2017co}, is the first successful attempt at supervised Deep Learning for LLP \cite{ardehaly2017co}. Inspired by consistency regularization in semi-supervision, LLP\_PI leveraged self-ensembling to produce consensus results under different regularization and augmentation mechanisms \cite{laine2016temporal}. LLP\_VAT employed Virtual Adversarial Training (VAT) to generate perturbed adversarial inputs, so as to conduct more robust label predictions \cite{tsai2020learning}. \citeauthor{liu2022llp} took advantage of Generative Adversarial Networks and proposed LLP-GAN to solve the LLP task \cite{liu2022llp}. 
% EasyLLP tried to convert LLP into a strongly supervised task by generating surrogate labels and then applied debiasing and variance reduction to minimize the corresponding errors introduced by surrogate labels \cite{busa2023easy}. \citeauthor{asanomi2023mixbag} proposed a novel bag-level data augmentation for LLP, with a confidence interval loss for the corresponding supervision \cite{asanomi2023mixbag}. 
LLP is a learning paradigm with special data settings. It was initially attempted to be solved using SVM \cite{yu2013proptosvm} and Logistic hypothesis \cite{yu2014learning}. With the development of deep learning, there have emerged many attempts with it for LLP tasks. Batch Averager is the first successful try to use supervised deep learning for LLP \cite{ardehaly2017co}. Inspired by consistency regularization in semi-supervision, LLP\_PI leveraged self-ensembling to produce consensus results under different regularization and augmentation mechanisms \cite{laine2016temporal}. LLP\_VAT exploits Virtual Adversarial Training to generate perturbed adversarial inputs to keep the a posteriori probabilities of the original and perturbed images consistent \cite{tsai2020learning}. LLPGAN introduces Generative Adversarial Network to perform instance-level classification by the discriminator, with the predicted probabilities for proportion losses \cite{liu2022llp}. LLPFC employs pseudo-labeling to each instance, based on the proportionally estimated noise transformation matrices \cite{zhang2022learning}. 
EasyLLP tried to convert LLP into a strongly supervised task by generating surrogate labels, with a debiasing and variance reduction to minimize the introduced errors \cite{busa2023easy}. 
MixBag is a bag-level data augmentation technique for LLP that introduces a loss of confidence intervals based on the proportion distributions in the sampled mixed bags \cite{asanomi2023mixbag}. 
\section{Methodology}
\label{sec:method}
Our goal is to achieve fine-grained representation learning without instance-level labeling through the LLP paradigm, so as to provide privacy protection for FGVC and reduce the risk of sensitive information leakage from the data collection source. 
We will introduce our Learning from Hierarchical Fine-Grained Label Proportions (LHFGLP) framework specifically designed for FGVC in \cref{model3} in four sections. First, we start with a problem formulation for the LLP settings (\cref{sec1}), followed by an overview of the entire architecture (\cref{sec2}). Then, we will give in-depth descriptions of our proposed Unrolled Hierarchical Fine-Grained Sparse Dictionary Learning strategy (\cref{sec3}) and Hierarchical Proportion Loss (\cref{sec4}) in LHFGLP. 

% \begin{figure*}[t]
% \centering
% % \fbox{\rule{0pt}{2in} \rule{0.9\linewidth}{0pt}}
% \includegraphics[width=0.8\textwidth]{models/model3.pdf} % Reduce the figure size so that it is slightly narrower than the column.
% \caption{Overview of our \textbf{UHFGSDL} framework.}
% \label{model3}
% \end{figure*}

%-------------------------------------------------------------------------
\subsection{Problem Formulation}
\label{sec1}
In LLP paradigm, datasets are given in image bags for model training, denoted as $\mathbf{B} = \{\mathbf{B}^1, ..., \mathbf{B}^i, ... \mathbf{B}^n\}$. Each bag $\mathbf{B}^i = \left\{ x_j^i \right\}_{j=1}^{\left| \mathbf{B}^i \right|}$ consists a collection of unlabeled image instances and is annotated with a bag-level label proportion $\mathbf{p}^i = (p_1^i, ..., p_c^i, ..., p_C^i)$, where $C$ is the total number of categories. Proportion $\mathbf{p}^i$ specifies that there are $p_c^i \times |\mathbf{B}^i|$ instances in the i-th bag $\mathbf{B}^i$ belonging to category $c \in \{1, ..., C\}$. 
For FGVC, all instances in the bags are fine-grained images, and our goal is to learn an instance-level fine-grained image classifier using only the bag-level annotations. Crucially, instance-level labels for each training image is inaccessible.
% Supervision only allows the use of bag-level label proportions, and we cannot access the category labels for each fine-grained image specifically.
%-------------------------------------------------------------------------
\subsection{Overview}
\label{sec2}
Within the LLP paradigm, we provide privacy protection with bag-level annotations to bypass direct instance-level supervision, but also confront challenges in capturing subtle distinctions across fine-grained categories. To address this shortcoming of conventional LLP methods, we propose
% LLP paradigm utilizes bag-level annotations for efficient model training without direct access to instance labels for privacy protection. For FGVC in LLP paradigm, we proposed 
a dedicated LLP framework called \textbf{Learning from Hierarchical Fine-Grained Label Proportions (LHFGLP)} to extract discriminative features for fine-grained subcategories, as illustrated in \cref{model3}. 
It adopts an \textbf{Unrolled Hierarchical Fine-Grained Sparse Dictionary Learning} strategy for fine-grained feature exploration.
% with progressive refinement of feature granularity. 
This strategy stems from three innovations in feature representation learning. By deep unrolling, traditional handcrafted iterative dictionary learning process can be transformed into learnable network optimization, corresponding to an \textit{unrolled dictionary learning} procedure. Thus, we could obtain an over-complete dictionary with \textit{sparsity constraints}, where features can be represented as sparse linear combinations of feature atoms, forcing the model focus on the most relevant discriminative regions. In addition, the unrolled dictionary learning process facilities progressive refinement of feature granularity through the inherent \textit{hierarchical structure} of \textit{fine-grained} categories. Furthermore, our proposed \textbf{Hierarchical Proportion Loss} will integrate with it to guide the progressive fine-grained feature refinement under bag-level supervision. 
% Our proposed Hierarchical Proportion Loss integrates with the feature refinement procedure and utilizes the inherent hierarchical structure of fine-grained categories for supervision.

% Following the and maintains a category-specific dictionary. 
% Utilizing the learned over-complete dictionary, images will be represented as linear combinations of dictionary atoms, capturing the essential and discriminative features for the downstream FGVC task.
% During the dictionary learning process, our proposed \textbf{Hierarchical Proportion Loss} will exploit the inherent hierarchical structure of fine-grained images to perform hierarchical supervision as the learning granularity progressively increases.
% During dictionary learning, the inherent hierarchical structure of fine-grained images is exploited to progressively increase the learning granularity, with the use of our proposed \textbf{Hierarchical Proportion Loss} for supervision.

%-------------------------------------------------------------------------
\subsection{Unrolled Hierarchical Fine-Grained Sparse Dictionary Learning}
\label{sec3}
\begin{figure}[t]
\centering
\includegraphics[width=\columnwidth]{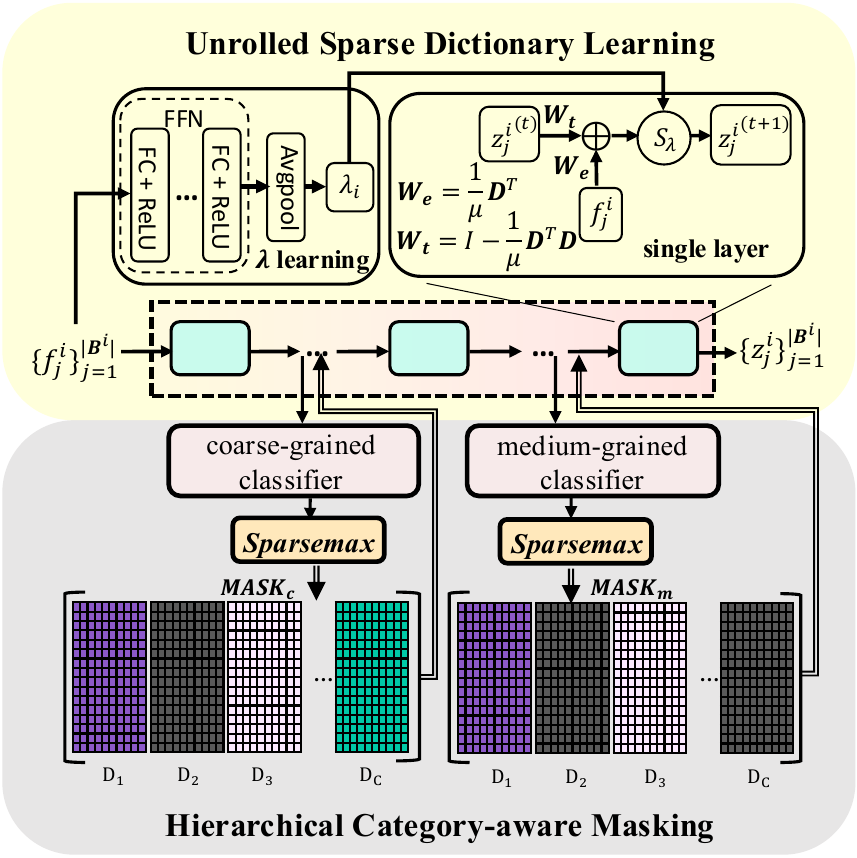} 
\caption{Implementing details of the Unrolled Hierarchical Fine-Grained Sparse Dictionary Learning strategy with Hierarchical Category-aware Masking.}
\label{model4}
\end{figure} 

In this section, we will explain our proposed Unrolled Hierarchical Fine-Grained Sparse Dictionary Learning strategy in detail from the following three components.
\paragraph{Unrolled Sparse Dictionary Learning}
In LHFGLP, we introduce an Unrolled Hierarchical Fine-Grained Sparse Dictionary Learning strategy to mine category-related fine-grained feature representations.
The optimization objective is based on the minimization of reconstruction error, written as:
\begin{equation}
\label{equation1}
\min_{\mathbf{D}, \mathbf{Z}} \sum_{i=1}^{n} \left( \frac{1}{2} \|\mathbf{F}^i - \mathbf{D} \mathbf{Z}^i\|_2^2 + \lambda \|\mathbf{Z}^i\|_1 \right),
\end{equation}
where $\mathbf{Z}^i=\{z_j^i\}_{j=1}^{|\mathbf{B}^i|}$ is the derived sparse representation of image bag $\mathbf{B}^i$ from its feature representation $\mathbf{F}^i=\{f_j^i\}_{j=1}^{|\mathbf{B}^i|}$ from Feature Extractor. The sparsity constraints are employed by $L1$ norm, and $\lambda$ is the regularization parameter to control sparsity.
However, as both the dictionary and sparse feature representations are unknown, the optimization problem becomes non-convex. Classical dictionary learning methods \cite{daubechies2004iterative, beck2009fast} address this issue in an alternating iterative manner, but are unable to be compatible with deep neural networks for end-to-end learnability.
% As pointed out in \cref{uDL}, the intuitive approach is to optimize iteratively \cite{daubechies2004iterative}. However, it is not compatible with deep neural networks for end-to-end training. 
Inspired by SC-MIL \cite{qiu2023sc}, we leverage the deep unrolling technique to unroll the handcrafted iterative optimization process into a cascade of L-layer single network layers, enabling it to be end-to-end trainable. The detailed implementation is illustrated in the upper part of \cref{model4}.

To carry out deep unrolling, we first re-parameterize the iterative optimization objective as:
\begin{equation}
\label{equation4}
\mathbf{Z}^{i,(t+1)} = S_{\lambda} (\mathbf{W_t}\mathbf{Z}^{i,(t)} + \mathbf{W_e}\mathbf{F}^i),
\end{equation}
\begin{equation}
\label{equation5} 
\mathbf{W_t} = I - \frac{1}{\mu}\mathbf{D}^T\mathbf{D},
\mathbf{W_e} = \frac{1}{\mu}\mathbf{D}^T,
\end{equation}
where $t \in \{1, ..., L\}$ denotes the $t$-th optimization iteration, and $S_{\lambda}(*) = \text{sign}(*) \cdot \max(|*|-\lambda, 0)$ is the element-wise soft-thresholding operator. The number of unrolled layers $L$ is a hyperparameter to be tuned for balancing the trade-off between model complexity and performance.
With the learning of dictionary $\mathbf{D}$, stepsize $\mu$, and sparse strength $\lambda$, each iterative update can be recast as a single network layer in \cref{model4}.
Following \cite{qiu2023sc}, an additional regression task is constructed to obtain the optimal sparse strength $\lambda_i$ for each image bag, corresponding to the $\lambda$ learning module. Also, the stepsize $\mu$ is another learnable global parameter for dictionary learning.
% Thus, the entire Dictionary Learning process is Deep Learning trainable, leading to more robust feature extraction.
% the stepsize $\mu$ is initialized to the square of the spectral norm of dictionary $\mathbf{D}$
Thus, we could obtain a more discriminative sparse feature representation of fine-grained image bags to facilitate subsequent fine-grained image classification.

\paragraph{Hierarchical Category-aware Masking}
In Unrolled Sparse Dictionary Learning, we aim to minimize the reconstruction error with the learned dictionary and sparse representations, which is often effective for signal repairing but neglects the category information for classification. In our work, the ultimate goal is to make the sparse representations as discriminative as possible, especially for visually similar fine-grained categories under the LLP paradigm.
To address this, we further design a Hierarchical Category-aware Masking scheme to integrate fine-grained hierarchy in the Unrolled Sparse Dictionary Learning procedure, enhancing with progressive refinement of discriminative feature representations. 
To include category information, we first re-represent our dictionary to be category-related, denoted as $\mathbf{D} = [\mathbf{D_1}, \mathbf{D_2}, ..., \mathbf{D_c}, ..., \mathbf{D_C}] = [d_1, ..., d_{n\_atoms}, ..., d_{c \times n\_atoms},..., d_{C \times n\_atoms}]$, where each $\mathbf{D_c}$ encodes the category-related features with $n\_atoms$ atoms.
During the unrolled dictionary learning procedure, we constructed classifiers at different granularities to generate dictionary masks (e.g., $\mathit{MASK_c}$ and $\mathit{MASK_m}$) that selectively suppress irrelevant category components in our category-related dictionary:
\begin{equation}
\mathbf{D}^{(t)} = \mathbf{D}^{(t)} \odot \mathit{MASK_i},
\end{equation}
where $\odot$ denotes element-wise masking and $i$ corresponds to the masking granularity in hierarchy.
Then, the masked dictionary will be used in subsequent unrolled dictionary learning steps. 
% Leveraging the inherent fine-grained hierarchical structure,
% in fine-grained image annotations, we construct multiple classifiers at different granularities during the unrolled dictionary learning procedure. From the outputs of these classifiers, masks are generated at the corresponding granularity to suppress the irrelevant category parts of the dictionary at each granularity, and then the masked dictionary is used for subsequent computations. 
% $\mathit{MASK_c}$ and $\mathit{MASK_m}$ in \cref{model4} indicate the generated masks at coarse and medium granularity in a three-level fine-grained hierarchy, which are used to mask the dictionary at the corresponding granularity.
% With the gradually increased granularity of dictionary masks, the learning granularity of our category-related dictionary is also progressively increased in a hierarchical manner, thus including more category information to achieve more discriminative category-related feature representations.
As the granularity of dictionary masks increases, the corresponding learning granularity of our category-related dictionary also increases progressively in a hierarchical manner. Therefore, the hierarchical category information is incorporated into the Unrolled Sparse Dictionary Learning procedure to achieve more discriminative category-related feature representations.
% enhancing the dictionary learning process by providing more category information.

\vspace{-10pt}
\paragraph{Sparsemax Activation}
\label{sparsemax}
Softmax is a prevalent activation function for image classification tasks to convert classifier logits into label probability distributions. The activated probability distribution is generally dense where even those categories with minimal relevance will retain non-zero probabilities. However, this may not be ideal for FGVC under the LLP paradigm, as we typically involves hundreds of visually similar subcategories and individual LLP bags seldom contain instances from all possible categories. 
% the number of fine-grained categories is quite large and the LLP bag generally does not contain images of all categories. 
In this case, Sparsemax\cite{martins2016softmax} is able to induce sparse probability distributions, nullifying probabilities for irrelevant categories within LLP bags to generate effective dictionary masks. 
% which means some probability values will be activated to zero and the corresponding categories will be assumed not to exist at all. 
% It is consistent with the implicit assumption of FGVC in the LLP paradigm that some categories do not appear in the bag with zero label proportions.
In addition, the sparse probability distributions also improve the discriminative power by directing our model to focus on fewer but more relevant categories.
Therefore, we exploit Sparsemax instead of the widely-used Softmax to obtain the probability distribution, filter those irrelevant categories to generate more robust dictionary masks for our Unrolled Sparse Dictionary Learning.

\subsection{Hierarchical Proportion Loss}
\label{sec4}
% \subsubsection{Proportion Loss}
Proportion Loss is the standard supervision method for LLP tasks \cite{ardehaly2017co}. Specifically, it is the cross-entropy between the ground-truth label proportion $\mathbf{p^i} = (p_1^i, ..., p_c^i, ..., p_C^i)^T$ and the estimated $\mathbf{\hat{p}^i} = (\hat{p}_1^i, ..., \hat{p}_c^i, ..., \hat{p}_C^i)^T$ of bag $B^i$,
% which is the average of predicted labels of all instances in this bag. The Proportion Loss is defined as:
defined as:
\begin{equation}
L_{prop} = - \sum_{c=1}^{C} p_c^i \log \hat{p}_c^i ,
\end{equation}
\begin{equation}
\hat{p}_c^i = \frac{1}{|\mathbf{B^i}|} \sum_{j=1}^{|\mathbf{B^i}|} f(x_j^i)_c,
\end{equation}
where $f(x_j^i)_c$ is the predicted probability that instance $x_j^i$ belongs to class $c$.
% \subsubsection{Hierarchical Proportion Loss}
As we have constructed classifiers at different granularities for dictionary masking, we will further take advantage of this to conduct hierarchical bag-level supervision to facilitate our progressive feature refinement.
% (e.g., $\hat{p}_{coarse}^i$, $\hat{p}_{medium}^i$, and $\hat{p}_{fine}^i$). 
Once we obtained the label proportion at fine-grained level, it is straightforward to get access to the label proportions at any coarser level of granularity.
% The hierarchical structure in FGVC enables us to obtain the label proportions at all hierarchies once we know it at any one of them. 
Thus, each image bag will be equipped with multiple label proportions corresponding to each level in the hierarchical structure, such as $p_{coarse}^i$, $p_{medium}^i$ and $p_{fine}^i$ (i.e., $p^i$).
% FGVC inherently maintains the hierarchical structure, so once the label and label proportion in any hierarchy is known, the corresponding labels and label proportions in other hierarchical levels are also available. Therefore, each bag will be annotated with multiple ground-truth label proportions such as $p_{coarse}^i$, $p_{medium}^i$ and $p_{fine}^i = p^i$ in three hierarchical levels. 
As a result, we design the Hierarchical Proportion Loss for our LHFGLP framework, tightly integrated with the Unrolled Hierarchical Fine-Grained Sparse Dictionary Learning strategy for progressive feature refinement, which is written as:
\begin{equation}
L_{h\_prop} = \sum_{l=1}^{H} L_{prop}^l, 
\end{equation}
where $L_{prop}^l$ denotes the Proportion Loss at level $l \in \{1, ..., H\}$ of the hierarchy, with $H$ hierarchies in total. 
\section{Experiments}
\label{sec:experiments}

\subsection{Experimental Setups}
\paragraph{Bag Preparation}
\label{bag_gen}
Our work comes from a completely new perspective in privacy-preserving FGVC by packing fine-grained images into bags to conduct accurate fine-grained image classification, without direct access to individual instance labels. As there is no readily available LLP dataset dedicated to FGVC, we first construct fine-grained image bags with bag-level annotations. Three widely used FGVC datasets, including \textbf{CUB-200-2011 (CUB})\cite{wah2011caltech}, \textbf{FGVC Aircraft (Aircraft)}\cite{maji2013fine}, and \textbf{Stanford Cars (Cars)}\cite{krause20133d} are used.
% to evaluate the LLP task dedicated to FGVC. % , with the hierarchical structure introduced by \citeauthor{chang2021your} \cite{chang2021your}.
To prepare image bags, a fixed number of images are randomly selected from the dataset each time to be packed, eliminating implied data and label distributions introduced by elaborate manual construction.
Then, the proportion of each category is calculated for each bag based on the published instance-level annotations. 
Meanwhile, the category labels of individual images will be hidden and only the label proportions will be released as available bag-level annotations.
In real-world applications, different bags do not typically carry the same instances, so we do not allow overlapped sampling of instances to construct image bags. To maintain the same experimental setup on all datasets, we keep the bag size to 10 for all our LLP datasets.

\vspace{-10pt}
\paragraph{Baselines}
As a new paradigm for FGVC with privacy preservation, we investigated a more dedicated LLP approach for FGVC. Therefore, we choose current mainstream LLP algorithms as our baselines,
including LLP \cite{ardehaly2017co}, LLP\_PI \cite{laine2016temporal}, LLP\_VAT \cite{tsai2020learning} and LLP+MixBag \cite{asanomi2023mixbag}.
Since these methods are mainly used to solve general-purpose image classification tasks, we further combine them with the same FGVC pipeline for better task adaptability and fair comparisons.
For extensive evaluation, we also provide state-of-the-art FGVC approaches with instance-level supervision, consisting of Vanilla ResNet-50, FGoN \cite{chang2021your}, LHT \cite{wang2021label}, PMG \cite{du2020fine} and MGCF \cite{shu2023fine}.

\vspace{-10pt}
\paragraph{Implementation Details}
% We adopted ResNet-50 pre-trained on ImageNet as the feature extraction backbone and resized each input image to $224 \times 224$ over the entire experiment. To ensure fairness, all approaches were trained in the same bag construction. The hierarchical structures for FGVC datasets follow \cite{chang2021your}. 
To ensure fairness, all baseline approaches and our proposed LHFGLP use the same backbone ResNet-50 for feature extraction. It is initialized with ImageNet pre-trained weights, while other model components are randomly initialized.
% We also conducted performance comparisons with the Transformer-based architecture. 
Since our LHFGLP can be integrated into any FGVC pipelines in a plug-and-play manner, it also supports the use of more advanced backbones for feature extraction, such as ViT \cite{dosovitskiy2020image} and CLIP \cite{radford2021learning}, etc..
Throughout all experiments, the resolution of each input image (or bag of images) is resized to $224 \times 224$, and all approaches are guaranteed to be trained on the same collection of image bags. The hierarchical structures for FGVC datasets follow the work in \cite{chang2021your}. 
Momentum SGD is chosen as the optimizer and cosine annealing
\cite{loshchilov2016sgdr} 
is used as the learning scheduler with an initial learning rate of 0.005, which gradually decays to 0 under 100 epochs. Our weight decay is decided to $5 \times 10^{-4}$, and the momentum value is 0.9. We also employed common data augmentation methods, such as horizontal flipping, random cropping.

\vspace{-10pt}
\paragraph{Evaluation Metrics}
% Since the ultimate goal of the LLP learning paradigm is to train instance-level classifiers, all results evaluated in this paper represent instance-level prediction accuracies. Each experiment for model evaluation is run three times with the mean and standard deviation of the results across three attempts being reported.
Since our ultimate goal remains to provide instance-level fine-grained image classification, all evaluation results in this paper are measured by instance-level prediction accuracy. In particular, three independent evaluation experiments are conducted for each model, with the mean and standard deviation from three trials reported.

%-------------------------------------------------------------------------

\subsection{Experimental Results}

\cref{tab:result1} gives the comparative results of our \textbf{LHFGLP} with other LLP approaches, and the best classification accuracies achieved under bag-level supervision are highlighted in bold.
Experimental results on all three LLP-based FGVC 
datasets show that our LHFGLP offers more powerful ability to extract fine-grained features, significantly improving the performance of fine-grained image classification. For instance, in comparison to LLP\_VAT, we achieved \textbf{2.53\%}, \textbf{4.83\%} and \textbf{3.46\%} improvement in classification accuracy on \textbf{CUB}, \textbf{Aircraft} and \textbf{Cars}, respectively. 
% The best classification accuracies achieved by these LLP algorithms under bag-level supervision are highlighted in bold.
% and the best performance under instance-level supervision is marked with underlines. 

In order to provide a more robust evaluation, we also give the current state-of-the-art instance-level supervised FGVC approaches for benchmarking in \cref{tab:result_instance}, with the best performance underlined. Results demonstrate that there is an expected performance gap between LLP-based approaches and instance-level methods due to the inherent information loss associated with privacy-preserving bag constraints. Considering the inaccessibility of instance labels, the performance achieved by our LHFGLP remains highly competitive, proving its feasibility for FGVC with privacy requirements.
% However, our method exhibits the optimal classification accuracy under bag-level supervision. 

\begin{table}[t]
\centering
\fontsize{9pt}{9pt}\selectfont
\resizebox{0.99\columnwidth}{!}{
\begin{tabular}{@{}lr|c|c|c@{}}
\toprule
\multicolumn{2}{c|}{\multirow{2}{*}{Methods}} & \multicolumn{3}{c}{Accuracy(\%)}  \\ 
\cmidrule(lr){3-5}
\multicolumn{2}{c|}{} & \textbf{CUB} & \textbf{Aircraft} & \textbf{Cars} \\ 
\midrule
\multicolumn{2}{c|}{LLP \cite{ardehaly2017co}} &  $73.31_{\pm 0.12}$ & $71.38_{\pm 0.16}$ & $73.40_{\pm 0.18}$  \\
\multicolumn{2}{c|}{LLP\_PI \cite{laine2017temporal}} & $73.59_{\pm 0.03}$ & $71.94_{\pm 0.13}$ & $73.38_{\pm 0.10}$ \\ 
\multicolumn{2}{c|}{LLP\_VAT \cite{tsai2020learning}} & $73.52_{\pm 0.05}$ & $71.85_{\pm 0.08}$ & $73.76_{\pm 0.05}$ \\ 
\multicolumn{2}{c|}{{LLP+MixBag \cite{asanomi2023mixbag}}} &  $72.95_{\pm 0.59}$ & $69.67_{\pm 0.24}$ & $68.56_{\pm 0.36}$ \\ 
\midrule
\multicolumn{2}{c|}{{\textbf{LHFGLP}}} & 
$\mathbf{76.05_{\pm 0.04}}$ & $\mathbf{76.68_{\pm 0.02}}$ & $\mathbf{77.22_{\pm 0.02}}$ \\
\bottomrule
\end{tabular}
}
\caption{Experimental results on three fine-grained datasets using mainstream LLP algorithms and our LHFGLP. All results are the means of three experimental trials with their standard deviations. The best results are highlighted in bold.}
\label{tab:result1}
\end{table}

\begin{table}[t]
\centering
\fontsize{8.5pt}{8.5pt}\selectfont
% \resizebox{0.99\columnwidth}{!}{
\begin{tabular}{@{}l|lr|c@{}}
\toprule
% \multirow{2}{*}{\rotatebox{90}{\textbf{\fontsize{6.5pt}{6.5pt}\selectfont Supervision}}} & \multicolumn{2}{c|}{\multirow{2}{*}{Methods}} & \multicolumn{3}{c}{Accuracy(\%)}  \\ 
& \multicolumn{2}{c|}{\multirow{2}{*}{Methods}} & Accuracy(\%)  \\ 
% \cmidrule(l){4}
\cmidrule(lr){4-4}
& \multicolumn{2}{c|}{} & CUB \\ 
\midrule
\multirow{5}{*}{\rotatebox{90}{\textbf{\fontsize{6.5pt}{6.5pt}\selectfont Instance-level\textsuperscript{*}}}} & \multicolumn{2}{c|}{Vanilla  ResNet-50} & $74.24$ \\
% & \multicolumn{2}{c|}{MC \cite{chang2020devil}} & $77.85$  \\
& \multicolumn{2}{c|}{FGoN \cite{chang2021your}} & $77.95$  \\
& \multicolumn{2}{c|}{LHT \cite{wang2021label}} & $79.29$  \\
% & \multicolumn{2}{c|}{NTS \cite{yang2018learning}}  & $80.45$ \\
% & \multicolumn{2}{c|}{CGVC \cite{chen2022cross}}  & $80.54$ \\
& \multicolumn{2}{c|}{PMG \cite{du2020fine}}  & $\underline{82.26}$ \\
& \multicolumn{2}{c|}{MGCF \cite{shu2023fine}}  & $81.68$ \\
\midrule
\multirow{6}{*}{\rotatebox{90}{\textbf{ \fontsize{6.5pt}{6.5pt}\selectfont Bag-level}}} & \multicolumn{2}{c|}{LLP \cite{ardehaly2017co}} &  $73.31_{\pm 0.12}$  \\
& \multicolumn{2}{c|}{LLP\_PI \cite{laine2017temporal}} & $73.59_{\pm 0.03}$ \\ 
& \multicolumn{2}{c|}{LLP\_VAT \cite{tsai2020learning}} & $73.52_{\pm 0.05}$ \\ 
& \multicolumn{2}{c|}{{LLP+MixBag \cite{asanomi2023mixbag}}} &  $72.95_{\pm 0.59}$ \\ 
% \midrule
\cmidrule(lr){2-4}
& \multicolumn{2}{c|}{{\textbf{LHFGLP}}} & 
$\mathbf{76.05_{\pm 0.04}}$  \\

\bottomrule
\end{tabular}
% }
\caption{Performance comparison of algorithms under instance-level vs. bag-level supervision on the CUB dataset. \textsuperscript{*} For fairness, the input resolution of approaches under instance-level supervision in this paper is consistent with our bag-level supervision settings, which is the same $224 \times 224$.}
\label{tab:result_instance}
\end{table}

\subsection{Ablation Study and Visualization Analysis}
% \begin{table*}[h]
% \centering
% \fontsize{10pt}{10pt}\selectfont
% % \resizebox{0.99\columnwidth}{!}{
% \begin{tabular}{@{}lr|ccccc|>{\centering\arraybackslash}p{2cm}@{}}
% \toprule
% \multicolumn{2}{c|}{\multirow{2}{*}{Methods}} & \multicolumn{5}{c|}{Components} & Accuracy(\%) \\ 
% \cmidrule(lr){3-7}
% \cmidrule(lr){8-8}
% \multicolumn{2}{c|}{} & USDL & $MASK_c$ & $MASK_m$ & Softmax & Sparsemax & Aircraft \\
% % \midrule
% % \multicolumn{2}{c|}{LLP} & & & & & & 71.38  \\
% % \multicolumn{2}{c|}{LLP\_PI} & & & & & & 71.94 \\ 
% % \multicolumn{2}{c|}{LLP\_VAT} & & & & & & 71.85 \\ 
% % \multicolumn{2}{c|}{LLP+MixBag} & & & & & & 69.67 \\ 
% \midrule
% \multicolumn{2}{c|}{\multirow{6}{*}{LHFGLP}} & $\times$ & $\times$ & $\times$ & - & - & 71.38\\
% \multicolumn{2}{c|}{} & \checkmark & $\times$ & $\times$ & - & - & 76.31\\
% \multicolumn{2}{c|}{} & \checkmark & \checkmark &  &  & \checkmark & 76.53\\
% \multicolumn{2}{c|}{} & \checkmark &  & \checkmark &  & \checkmark & 76.56\\
% \multicolumn{2}{c|}{} & \checkmark & \checkmark & \checkmark & \checkmark &  & 76.47\\
% \multicolumn{2}{c|}{} & \checkmark & \checkmark & \checkmark &  & \checkmark & \textbf{76.68}\\
% \bottomrule
% \end{tabular}
% % }
% \caption{Ablation results using different granularity of hierarchical masks and proportion loss.}
% \label{tab:ab1}
% \end{table*}

\paragraph{Effectiveness of Unrolled Sparse Dictionary Learning}
Our LHFGLP can be seamlessly integrated into existing FGVC pipelines for privacy-sensitive applications in a modular and plug-and-play fashion with Unrolled Hierarchical Fine-Grained Sparse Dictionary Learning. To demonstrate our effectiveness, we experimentally compare the performance gains achieved by this strategy. Without the use of our proposed module, the entire framework would degrade into an basic LLP learning process based solely on the fundamental FGVC pipeline. Ablation results are given in \cref{tab:ab}, which demonstrates the benefits of using our approach with 2.74\%, 5.30\% and 3.82\% performance gain on CUB, Aircraft and Cars, respectively.
\begin{table}[t]
\centering
\fontsize{9pt}{9pt}\selectfont
\resizebox{0.99\columnwidth}{!}{
\begin{tabular}{@{}lr|>{\centering\arraybackslash}p{1cm}|>{\centering\arraybackslash}p{1cm}|>{\centering\arraybackslash}p{1.1cm}@{}}
\toprule
\multicolumn{2}{c|}{\multirow{2}{*}{Methods}} & \multicolumn{3}{c}{Accuracy(\%)}  \\ 
\cmidrule(lr){3-5}
\multicolumn{2}{c|}{} & \textbf{CUB} & \textbf{Aircraft} & \textbf{Cars} \\ 
\midrule
\multicolumn{2}{c|}{LHFGLP\textsubscript{without Dictionary Learning}} & 73.31 & 71.38 & 73.40 \\
\multicolumn{2}{c|}{LHFGLP\textsubscript{with Dictionary Learning} (\textbf{Ours})} & 
\textbf{76.05} & \textbf{76.68} & \textbf{77.22} \\
\bottomrule
\end{tabular}
}
\caption{Ablation results on whether using our Unrolled Hierarchical Fine-Grained Sparse Dictionary Learning.}
\label{tab:ab}
\end{table}

\vspace{-10pt}
\paragraph{Effectiveness of Hierarchical Category-aware Masking with Hierarchical Proportion Loss}
% In this section, we conduct ablation studies to compare the effect of using masks of different granularity to demonstrate the benefits of progressively increasing the learning granularity in our unrolled dictionary learning process. Ablation results on Aircraft are displayed in \cref{tab:ab1}. We can observe that our LHFGLP has achieved at least \textbf{4.37\%} accuracy improvement for fine-grained image classification even without any masks. Masking at coarse or medium granularity could provide additional category information and thus improve the classification accuracy compared to not performing any masking at all. Our LHFGLP gradually masks dictionary learning at multiple granularities and achieves optimal classification performance.
Our LHFGLP employs Hierarchical Category-aware Masking to progressively refine fine-grained feature learning with the corresponding level of hierarchical supervision via Hierarchical Proportion Loss. As our Hierarchical Proportion Loss takes advantage of the classifiers constructed in Hierarchical Category-aware Masking for generating different granularities of masks, the ablation analysis of masks at different granularities is equivalent to the ablation assessment with different hierarchical levels of proportion loss.
To demonstrate their effectiveness in Unrolled Sparse Dictionary Learning, we conducted the following ablation studies. 
% To demonstrate the effectiveness of our progressive refinement of fine-grained features with Hierarchical Category-aware Masking during the Unrolled Sparse Dictionary Learning procedure, we conducted ablation studies to compare the performance of using different granularity masks or not. 
Ablation results on Aircraft are shown in \cref{tab:ab1}. 

We first evaluated the effectiveness of not using any masks, which corresponds to the Hierarchical Proportion Loss with only one layer of fine-grained bag supervision. Then we tested using only the coarse-grained mask $MASK_c$ or medium-grained mask $MASK_m$, with the addition of bag-level supervision at their corresponding granularity in the Hierarchical Proportion Loss. Our approach in LHFGLP utilizes masks of all granularities in a progressive manner during Unrolled Dictionary Learning, thus adopts the complete usage of our Hierarchical Proportion Loss design. The ablation results show that our LHFGLP improves the accuracy of at least 4.37\% on Aircraft, even without using any masks. On this basis, using coarse or medium granularity masks with corresponding hierarchical supervision can further provide category information to assist dictionary as well as feature representation learning, thus improving fine-grained classification accuracy. For our LHFGLP, the optimal classification performance is achieved through different granularities of masks and supervision, in cooperation with our Unrolled Sparse Dictionary Learning procedure for progressive fine-grained feature refinement.
\begin{table}[t]
\centering
\fontsize{8pt}{8pt}\selectfont
% \resizebox{0.99\columnwidth}{!}{
\begin{tabular}{@{}lr|>{\centering\arraybackslash}p{2cm}@{}}
\toprule
\multicolumn{2}{c|}{Methods} & \textbf{Aircraft} \\ 
\midrule
\multicolumn{2}{c|}{LLP} & 71.38  \\
\multicolumn{2}{c|}{LLP\_PI} & 71.94 \\ 
\multicolumn{2}{c|}{LLP\_VAT} & 71.85 \\ 
\multicolumn{2}{c|}{LLP+MixBag} & 69.67 \\ 
% \multicolumn{2}{c|}{LLP\_LargeBags} & 64.70 \\ 
\midrule
% \multicolumn{2}{c|}{\textbf{UHFGSDL\textsubscript{noMASK}}} & 76.31\\
% \multicolumn{2}{c|}{\textbf{UHFGSDL\textsubscript{MASK\textsubscript{c}}}} & 0.\\
% \multicolumn{2}{c|}{\textbf{UHFGSDL\textsubscript{MASK\textsubscript{m}}}} & 76.56 \\
\multicolumn{2}{c|}{LHFGLP\textsubscript{\textbf{noMASK}}} & 76.31\\
\multicolumn{2}{c|}{LHFGLP\textsubscript{\textbf{MASK\textsubscript{c}}}} & 76.53\\
\multicolumn{2}{c|}{LHFGLP\textsubscript{\textbf{MASK\textsubscript{m}}}} & 76.56 \\
% \multicolumn{2}{c|}{\textbf{LHFGLP\textsubscript{MASK\textsubscript{c}+MASK\textsubscript{m}}}} & \textbf{76.64}  \\
\multicolumn{2}{c|}{LHFGLP\textsubscript{MASK\textsubscript{c}+MASK\textsubscript{m}} (\textbf{Ours})} & \textbf{76.68}  \\
\bottomrule
\end{tabular}
% }
\caption{Ablation results using different granularity of hierarchical masks and proportion loss.}
\label{tab:ab1}
\end{table}

\vspace{-10pt}
\paragraph{Softmax vs. Sparsemax}
\begin{table}[t]
\centering
\fontsize{8pt}{8pt}\selectfont
% \resizebox{0.99\textwidth}{!}{
\begin{tabular}{@{}lr|>{\centering\arraybackslash}p{1cm}|>{\centering\arraybackslash}p{1cm}|>{\centering\arraybackslash}p{1.1cm}@{}}
\toprule
\multicolumn{2}{c|}{\multirow{2}{*}{Methods}} & \multicolumn{3}{c}{Accuracy(\%)}  \\ 
\cmidrule(lr){3-5}
\multicolumn{2}{c|}{} & \textbf{CUB} & \textbf{Aircraft} & \textbf{Cars} \\ 
\midrule
\multicolumn{2}{c|}{LLP} & 73.31 & 71.38 & 73.40 \\
\multicolumn{2}{c|}{LLP\_PI} & 73.59 & 71.94 & 73.38 \\ 
\multicolumn{2}{c|}{LLP\_VAT} & 73.52 & 71.85 & 73.76 \\ 
\multicolumn{2}{c|}{LLP+MixBag} & 72.95 & 69.67 & 68.56 \\ 
% \multicolumn{2}{c|}{LLP\_LargeBags} &  & 64.70 &  \\ 
\midrule
\multicolumn{2}{c|}{LHFGLP\textsubscript{\textbf{Softmax}}} & 75.48
 & 76.47 & 76.81 \\
% \multicolumn{2}{c|}{\textbf{LHFGLP\textsubscript{Sparsemax}}} & 
% \textbf{76.01} & \textbf{76.64} & \textbf{77.24} \\
\multicolumn{2}{c|}{LHFGLP\textsubscript{Sparsemax} (\textbf{Ours})} & 
\textbf{76.05} & \textbf{76.68} & \textbf{77.22} \\
\bottomrule
\end{tabular}
\caption{Ablation results using Softmax vs. Sparsemax activation for hierarchical dictionary masking generation.}
\label{tab:ab2}
\end{table}
% We also found that direct normalization of the classifier output using Softmax will distract the attention to each fine-grained category, resulting in the failure of masking to provide valid category information for guiding dictionary learning. 
% Table \ref{tab:ab2} displays the ablation results on all three fine-grained LLP datasets using Softmax or Sparsemax to normalize the classifier outputs for generating masks and subsequent masking. It is obvious that Softmax... \textbf{TODO}
As we have pointed out in \cref{sparsemax}, activating classifier outputs using Softmax will assign a probability score to each fine-grained category, even if its logits is very small. This approach implicitly invalidates our masking scheme, since no categories will be masked during Unrolled Dictionary Learning, thus distracting the attention from every fine-grained category. \cref{tab:ab2} displays the ablation results for three fine-grained LLP datasets with hierarchical masks generated using either Softmax or Sparsemax activated classifier outputs. 
% the guidance classifiers activated using Softmax or Sparsemax for hierarchical masks generation. 
We can observe that the masks obtained from Softmax-guided dictionary learning are slightly less accurate than those generated by Sparsemax, which demonstrates the effectiveness of our Sparsemax design.

\begin{figure}[t]
\centering
\includegraphics[width=1.05\columnwidth]{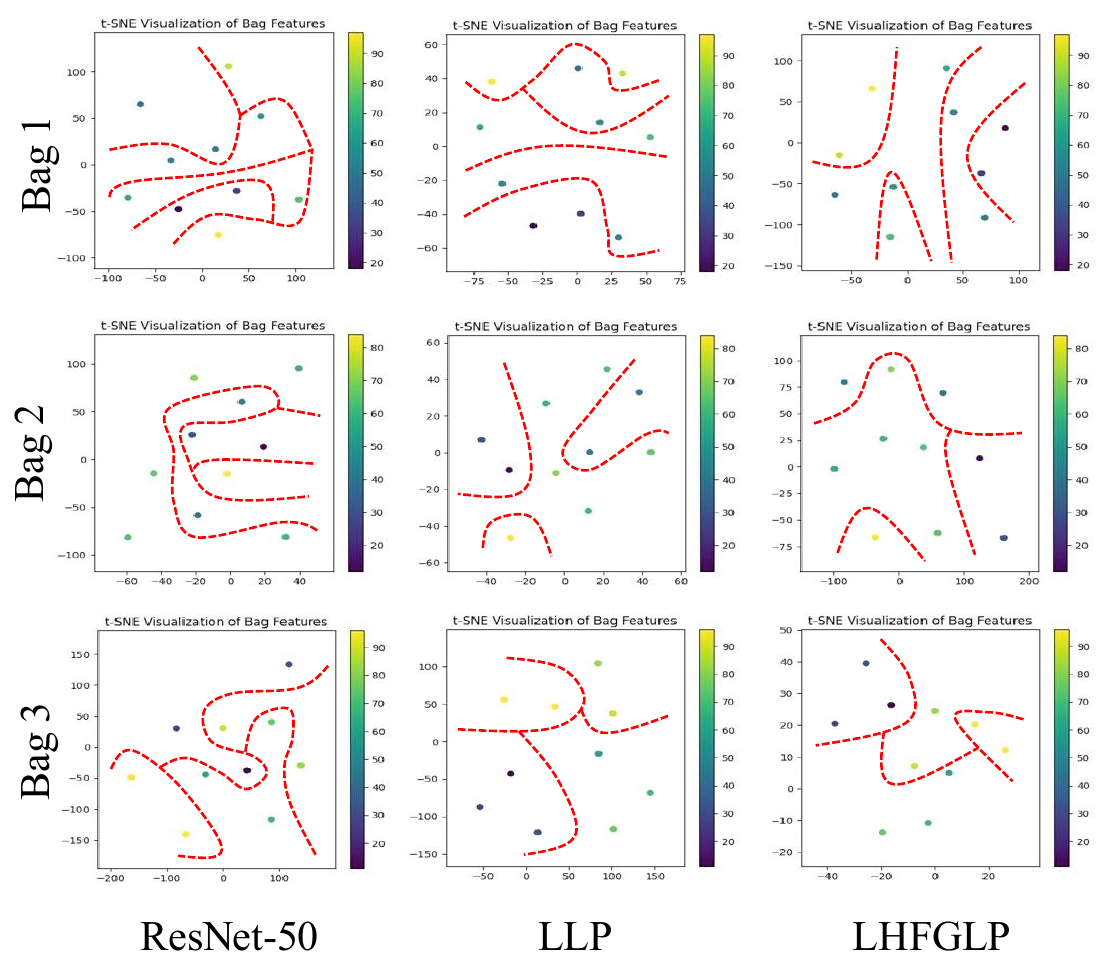} 
\caption{t-SNE visualization of three fine-grained image example bags (in row). 
Each one corresponds to a feature space obtained from \textbf{Vanilla ResNet-50} (left), baseline \textbf{LLP} (middle), and our \textbf{LHFGLP} (right). Colored points refer to the feature representation of each image. The red dashed lines indicate the efforts of classifiers to make the fine-grained distinctions.}
\label{tSNE}
\end{figure}

\vspace{-10pt}
\paragraph{Bag visualization}
The special bag-level data and annotations in LLP paradigm make it more likely to confuse instances within the same bag. Also, FGVC requires the ability to capture subtle intra-class distinctions between visually similar subcategories. Therefore, effective separation of different subcategories of instances, especially in the same image bag, is a visual indication on whether the extracted features are sufficiently discriminative or not.
% Effective separation of different classes of instances in the same bag can indicate whether the extracted features are discriminative enough.
Therefore, we use t-SNE \cite{van2008visualizing} to visualize the feature embeddings of instances in the same bag. \cref{tSNE} presents the feature space for three example bags, where each row corresponds to a fine-grained image bag. 
The leftmost column gives the feature spaces obtained by applying Vanilla ResNet-50 to fine-grained image bags, the middle column is derived from the classical LLP baseline \cite{ardehaly2017co}, and the rightmost column corresponds to our LHFGLP. 
As expected, FGVC under the LLP paradigm has a more chaotic feature space and more intricate decision boundaries, which can be demonstrated by the feature space visualization from Vanilla ResNet-50. The feature representations of different categories of instances are interspersed with each other, requiring tremendous efforts to distinguish between them. 
% This can be demonstrated in the feature space visualization extracted by the vanilla ResNet-50 architecture. 
In contrast, LLP extracts some discriminative features with clear classification boundaries, reducing the difficulty of fine-grained subtle classification. Our LHFGLP exhibits enhanced discriminative power by generating more separated feature representations and clearer decision boundaries.

\begin{figure}[t]
\centering
\includegraphics[width=0.91\columnwidth]{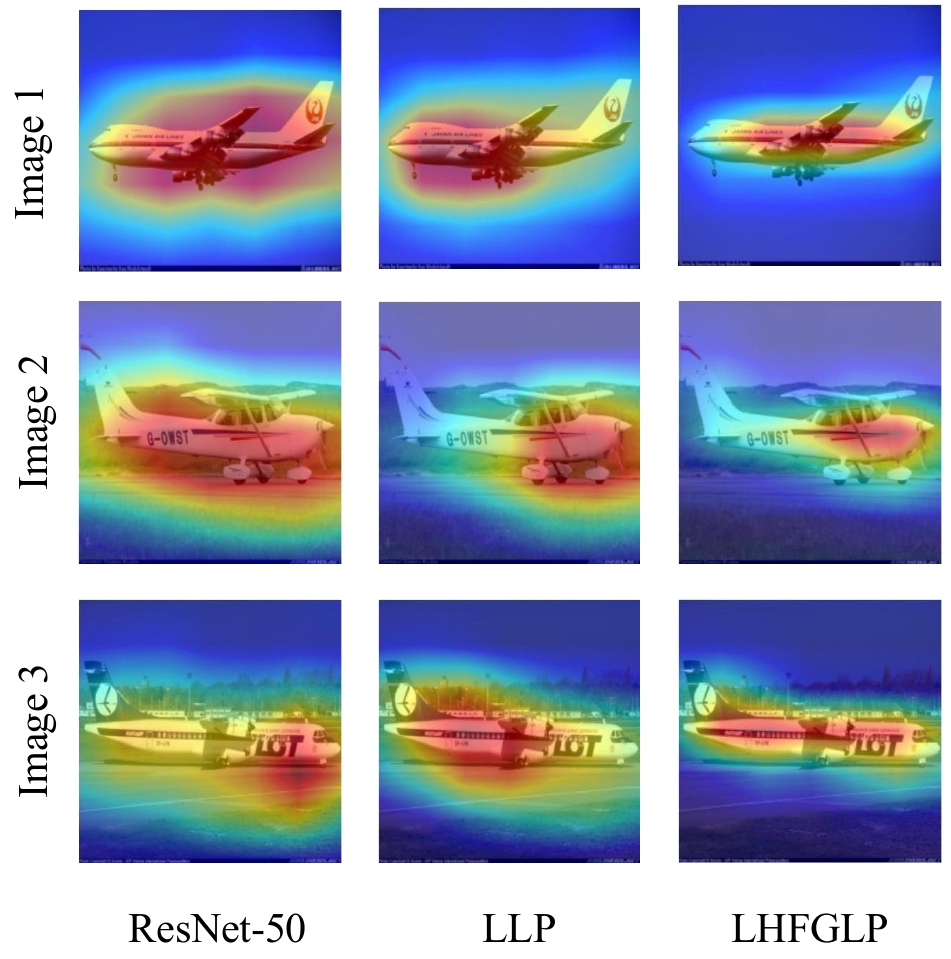} 
\caption{Activation visualization of three fine-grained images in the Aircraft dataset derived from \textbf{Vanilla ResNet-50} (left), baseline \textbf{LLP} (middle), and our \textbf{LHFGLP} (right). The highlighted parts refer to the supporting visual regions where the model is focusing its attention on for fine-grained image classification.}
\label{feature}
\end{figure}

\vspace{-10pt}
\paragraph{Feature visualization}
To further demonstrate the advantages of our proposed approach, we also conducted visualization analysis on the extracted features across different framework, including Vanilla ResNet-50, LLP and our LHFGLP. Each row in \cref{feature} corresponds to one fine-grained image in the Aircraft dataset. 
% For FGVC tasks, it is essential for models to pay more attention on local discriminative and detailed features. 
% We can observe that the feature activation of Vanilla ResNet-50 is spread across the regions of global contours of targets, with weaker response to features with local discriminative details. 
% In contrast, LLP is able to narrow the region of response towards significant sub-regions of the corresponding target. 
% Furthermore, our proposed LHFGLP exhibits superior local feature localization capability and thus achieves better concentration on those dominant discriminative regions.
In the context of FGVC, successful recognition critically depends on a model's capacity to focus on subtle but discriminative local characteristics, particularly when dealing with highly similar subcategories that primarily differ in specific part-level details. Our visualization results in \cref{feature} reveal distinct behavioral patterns among different approaches. The Vanilla ResNet-50 architecture, while demonstrating reasonable global feature recognition capabilities through its activation spread across target contours, shows limited sensitivity to localized discriminative features.
In comparison, the baseline LLP framework presents improved feature localization, concentrating activation responses towards significant sub-regions of the corresponding target. 
As for our proposed LHFGLP framework, it exhibits superior local feature localization capability and better concentration on dominant discriminative regions, thus achieves better fine-grained image classification accuracy.
The comparative visualization analysis suggests that while all these methods can identify the general object category, our LHFGLP is able to consistently locate and concentrate on the subtle and discriminative feature regions for fine-grained recognition. Especially with the weak bag-level supervision in the LLP paradigm, it becomes even more critical to realize accurate and effective fine-grained feature localization and representation. 

\section{Conclusion}
In this paper, we addressed the privacy concerns in fine-grained visual classification by leveraging the Learning from Label Proportions paradigm and the hierarchical structure of fine-grained datasets. We proposed LHFGLP, a framework that effectively learns instance-level fine-grained representations from bag-level data, without requiring direct access to instance labels. Our method demonstrates that fine-grained visual classification can be achieved under privacy constraints while maintaining high recognition accuracy.
In future work, we aim to extend our framework to other fine-grained tasks, such as fine-grained image retrieval. Additionally, we plan to investigate its adaptability to cross-modal learning and explore more efficient LLP-based learning strategies to further enhance performance in privacy-sensitive applications. 
{
    \small
    \bibliographystyle{ieeenat_fullname}
    \bibliography{main}

\begin{thebibliography}{33}
\providecommand{\natexlab}[1]{#1}
\providecommand{\url}[1]{\texttt{#1}}
\expandafter\ifx\csname urlstyle\endcsname\relax
  \providecommand{\doi}[1]{doi: #1}\else
  \providecommand{\doi}{doi: \begingroup \urlstyle{rm}\Url}\fi

\bibitem[Ardehaly and Culotta(2017)]{ardehaly2017co}
Ehsan~Mohammady Ardehaly and Aron Culotta.
\newblock Co-training for demographic classification using deep learning from label proportions.
\newblock In \emph{ICDMW}, 2017.

\bibitem[Asanomi et~al.(2023)Asanomi, Matsuo, Suehiro, and Bise]{asanomi2023mixbag}
Takanori Asanomi, Shinnosuke Matsuo, Daiki Suehiro, and Ryoma Bise.
\newblock Mixbag: Bag-level data augmentation for learning from label proportions.
\newblock In \emph{ICCV}, 2023.

\bibitem[Beck and Teboulle(2009)]{beck2009fast}
Amir Beck and Marc Teboulle.
\newblock A fast iterative shrinkage-thresholding algorithm for linear inverse problems.
\newblock \emph{SIAM journal on imaging sciences}, 2009.

\bibitem[Busa-Fekete et~al.(2023)Busa-Fekete, Choi, Dick, Gentile, and Medina]{busa2023easy}
Robert Busa-Fekete, Heejin Choi, Travis Dick, Claudio Gentile, and Andres~Munoz Medina.
\newblock Easy learning from label proportions.
\newblock In \emph{NIPS}, 2023.

\bibitem[Chang et~al.(2021)Chang, Pang, Zheng, Ma, Song, and Guo]{chang2021your}
Dongliang Chang, Kaiyue Pang, Yixiao Zheng, Zhanyu Ma, Yi-Zhe Song, and Jun Guo.
\newblock Your" flamingo" is my" bird": Fine-grained, or not.
\newblock In \emph{CVPR}, 2021.

\bibitem[Chen et~al.(2024)Chen, Zhang, Liu, An, Gao, and Qiu]{chen2024fet}
Huazhen Chen, Haimiao Zhang, Chang Liu, Jianpeng An, Zhongke Gao, and Jun Qiu.
\newblock Fet-fgvc: Feature-enhanced transformer for fine-grained visual classification.
\newblock \emph{Pattern Recognition}, 2024.

\bibitem[Daubechies et~al.(2004)Daubechies, Defrise, and De~Mol]{daubechies2004iterative}
Ingrid Daubechies, Michel Defrise, and Christine De~Mol.
\newblock An iterative thresholding algorithm for linear inverse problems with a sparsity constraint.
\newblock \emph{Communications on Pure and Applied Mathematics: A Journal Issued by the Courant Institute of Mathematical Sciences}, 2004.

\bibitem[Dosovitskiy et~al.(2020)Dosovitskiy, Beyer, Kolesnikov, Weissenborn, Zhai, Unterthiner, Dehghani, Minderer, Heigold, Gelly, et~al.]{dosovitskiy2020image}
Alexey Dosovitskiy, Lucas Beyer, Alexander Kolesnikov, Dirk Weissenborn, Xiaohua Zhai, Thomas Unterthiner, Mostafa Dehghani, Matthias Minderer, Georg Heigold, Sylvain Gelly, et~al.
\newblock An image is worth 16x16 words: Transformers for image recognition at scale.
\newblock \emph{ICLR}, 2020.

\bibitem[Du et~al.(2020)Du, Chang, Bhunia, Xie, Ma, Song, and Guo]{du2020fine}
Ruoyi Du, Dongliang Chang, Ayan~Kumar Bhunia, Jiyang Xie, Zhanyu Ma, Yi-Zhe Song, and Jun Guo.
\newblock Fine-grained visual classification via progressive multi-granularity training of jigsaw patches.
\newblock In \emph{ECCV}, 2020.

\bibitem[Gwilliam et~al.(2021)Gwilliam, Teuscher, Anderson, and Farrell]{gwilliam2021fair}
Matthew Gwilliam, Adam Teuscher, Connor Anderson, and Ryan Farrell.
\newblock Fair comparison: Quantifying variance in results for fine-grained visual categorization.
\newblock In \emph{WACV}, 2021.

\bibitem[He et~al.(2022)He, Chen, Liu, Kortylewski, Yang, Bai, and Wang]{he2022transfg}
Ju He, Jie-Neng Chen, Shuai Liu, Adam Kortylewski, Cheng Yang, Yutong Bai, and Changhu Wang.
\newblock Transfg: A transformer architecture for fine-grained recognition.
\newblock In \emph{AAAI}, 2022.

\bibitem[Ke et~al.(2023)Ke, Cai, Chen, Liu, and Guo]{ke2023granularity}
Xiao Ke, Yuhang Cai, Baitao Chen, Hao Liu, and Wenzhong Guo.
\newblock Granularity-aware distillation and structure modeling region proposal network for fine-grained image classification.
\newblock \emph{Pattern Recognition}, 2023.

\bibitem[Krause et~al.(2013)Krause, Stark, Deng, and Fei-Fei]{krause20133d}
Jonathan Krause, Michael Stark, Jia Deng, and Li Fei-Fei.
\newblock 3d object representations for fine-grained categorization.
\newblock In \emph{ICCVW}, 2013.

\bibitem[Laine and Aila(2016)]{laine2016temporal}
Samuli Laine and Timo Aila.
\newblock Temporal ensembling for semi-supervised learning.
\newblock \emph{arXiv preprint arXiv:1610.02242}, 2016.

\bibitem[Laine and Aila(2017)]{laine2017temporal}
Samuli Laine and Timo Aila.
\newblock Temporal ensembling for semi-supervised learning.
\newblock In \emph{ICLR}, 2017.

\bibitem[Liu et~al.(2022)Liu, Wang, Hang, Wang, Qi, Tian, and Shi]{liu2022llp}
Jiabin Liu, Bo Wang, Hanyuan Hang, Huadong Wang, Zhiquan Qi, Yingjie Tian, and Yong Shi.
\newblock Llp-gan: a gan-based algorithm for learning from label proportions.
\newblock \emph{IEEE transactions on neural networks and learning systems}, 2022.

\bibitem[Loshchilov and Hutter(2016)]{loshchilov2016sgdr}
Ilya Loshchilov and Frank Hutter.
\newblock Sgdr: Stochastic gradient descent with warm restarts.
\newblock \emph{arXiv preprint arXiv:1608.03983}, 2016.

\bibitem[Maji et~al.(2013)Maji, Rahtu, Kannala, Blaschko, and Vedaldi]{maji2013fine}
Subhransu Maji, Esa Rahtu, Juho Kannala, Matthew Blaschko, and Andrea Vedaldi.
\newblock Fine-grained visual classification of aircraft.
\newblock \emph{arXiv preprint arXiv:1306.5151}, 2013.

\bibitem[Martins and Astudillo(2016)]{martins2016softmax}
Andre Martins and Ramon Astudillo.
\newblock From softmax to sparsemax: A sparse model of attention and multi-label classification.
\newblock In \emph{ICML}, 2016.

\bibitem[Pan et~al.(2023)Pan, Yu, Zhang, and Gao]{pan2023ssfe}
Zicheng Pan, Xiaohan Yu, Miaohua Zhang, and Yongsheng Gao.
\newblock Ssfe-net: Self-supervised feature enhancement for ultra-fine-grained few-shot class incremental learning.
\newblock In \emph{WACV}, 2023.

\bibitem[Qiu et~al.(2023)Qiu, Xiao, Zhu, Wang, and Sotiras]{qiu2023sc}
Peijie Qiu, Pan Xiao, Wenhui Zhu, Yalin Wang, and Aristeidis Sotiras.
\newblock Sc-mil: Sparsely coded multiple instance learning for whole slide image classification.
\newblock \emph{arXiv preprint arXiv:2311.00048}, 2023.

\bibitem[Radford et~al.(2021)Radford, Kim, Hallacy, Ramesh, Goh, Agarwal, Sastry, Askell, Mishkin, Clark, et~al.]{radford2021learning}
Alec Radford, Jong~Wook Kim, Chris Hallacy, Aditya Ramesh, Gabriel Goh, Sandhini Agarwal, Girish Sastry, Amanda Askell, Pamela Mishkin, Jack Clark, et~al.
\newblock Learning transferable visual models from natural language supervision.
\newblock In \emph{ICML}, 2021.

\bibitem[Shu et~al.(2023)Shu, Zhang, Wang, Wang, and Yi]{shu2023fine}
Xin Shu, Lei Zhang, Zizhou Wang, Lituan Wang, and Zhang Yi.
\newblock Fine-grained recognition: Multi-granularity labels and category similarity matrix.
\newblock \emph{Knowledge-Based Systems}, 2023.

\bibitem[Sikdar et~al.(2024)Sikdar, Liu, Kedarisetty, Zhao, Ahmed, and Behera]{sikdar2024interweaving}
Arindam Sikdar, Yonghuai Liu, Siddhardha Kedarisetty, Yitian Zhao, Amr Ahmed, and Ardhendu Behera.
\newblock Interweaving insights: high-order feature interaction for fine-grained visual recognition.
\newblock \emph{International Journal of Computer Vision}, 2024.

\bibitem[Tsai and Lin(2020)]{tsai2020learning}
Kuen-Han Tsai and Hsuan-Tien Lin.
\newblock Learning from label proportions with consistency regularization.
\newblock In \emph{ACML}, 2020.

\bibitem[Van~der Maaten and Hinton(2008)]{van2008visualizing}
Laurens Van~der Maaten and Geoffrey Hinton.
\newblock Visualizing data using t-sne.
\newblock \emph{Journal of machine learning research}, 2008.

\bibitem[Wah et~al.(2011)Wah, Branson, Welinder, Perona, and Belongie]{wah2011caltech}
Catherine Wah, Steve Branson, Peter Welinder, Pietro Perona, and Serge Belongie.
\newblock The caltech-ucsd birds-200-2011 dataset.
\newblock 2011.

\bibitem[Wang et~al.(2021)Wang, Xiao, Jia, Han, Meng, et~al.]{wang2021label}
Renzhen Wang, Kaiwen Xiao, Xixi Jia, Xiao Han, Deyu Meng, et~al.
\newblock Label hierarchy transition: Delving into class hierarchies to enhance deep classifiers.
\newblock \emph{arXiv preprint arXiv:2112.02353}, 2021.

\bibitem[Wei et~al.(2018)Wei, Xie, Wu, and Shen]{wei2018mask}
Xiu-Shen Wei, Chen-Wei Xie, Jianxin Wu, and Chunhua Shen.
\newblock Mask-cnn: Localizing parts and selecting descriptors for fine-grained bird species categorization.
\newblock \emph{Pattern Recognition}, 2018.

\bibitem[Xu et~al.(2024)Xu, Li, Wang, Jiang, and Tang]{xu2024context}
Qin Xu, Sitong Li, Jiahui Wang, Bo Jiang, and Jinhui Tang.
\newblock Context-semantic quality awareness network for fine-grained visual categorization.
\newblock \emph{arXiv preprint arXiv:2403.10298}, 2024.

\bibitem[Yu et~al.(2013)Yu, Liu, Kumar, Tony, and Chang]{yu2013proptosvm}
Felix Yu, Dong Liu, Sanjiv Kumar, Jebara Tony, and Shih-Fu Chang.
\newblock $\backslash$proptosvm for learning with label proportions.
\newblock In \emph{ICML}, 2013.

\bibitem[Yu et~al.(2014)Yu, Choromanski, Kumar, Jebara, and Chang]{yu2014learning}
Felix~X Yu, Krzysztof Choromanski, Sanjiv Kumar, Tony Jebara, and Shih-Fu Chang.
\newblock On learning from label proportions.
\newblock \emph{arXiv preprint arXiv:1402.5902}, 2014.

\bibitem[Zhang et~al.(2022)Zhang, Wang, and Scott]{zhang2022learning}
Jianxin Zhang, Yutong Wang, and Clay Scott.
\newblock Learning from label proportions by learning with label noise.
\newblock \emph{NIPS}, 2022.

\end{thebibliography}
}
\end{document}